\documentclass[preprint,12pt]{elsarticle}
\usepackage{lineno,hyperref}
\usepackage{tabularx}
\usepackage{ulem}
\usepackage{xcolor}
\usepackage{booktabs}
\usepackage{multirow}
\usepackage{amsmath,amssymb,amsfonts}

\journal{Journal of Information Security and Applications}
\begin{document}

\begin{frontmatter}



\title{Image Forgery Localization via Guided Noise and Multi-Scale Feature Aggregation} 


\author[1]{Yakun Niu}
\ead{ykniu@henu.edu.cn}

\author[1]{Pei Chen}
\ead{chenpei@henu.edu.cn}

\author[1]{Lei Zhang\corref{cor1}}  
\ead{zhanglei@henu.edu.cn}

\author[1]{Lei Tan}
\ead{tanlei@henu.edu.cn}

\author[1]{Yingjian Chen}
\ead{yingjianchen@henu.edu.cn}

\affiliation[1]{organization={Henan University},
	addressline={School of Computer and Information Engineering}, 
	city={Kaifeng},
	postcode={475000}, 
	state={Henan},
	country={China}}


\cortext[cor1]{Corresponding author}

\begin{abstract}
Image Forgery Localization (IFL) technology aims to detect and locate the forged areas in an image, which is very important in the field of digital forensics. However, existing IFL methods suffer from feature degradation during training using multi-layer convolutions or the self-attention mechanism, and perform poorly in detecting small forged regions and in robustness against post-processing. To tackle these, we propose a guided and multi-scale feature aggregated network for IFL. Spectifically, in order to comprehensively learn the noise feature under different types of forgery, we develop an effective noise extraction module in a guided way. Then, we design a Feature Aggregation Module (FAM) that uses dynamic convolution to adaptively aggregate RGB and noise features over multiple scales. Moreover, we propose an Atrous Residual Pyramid Module (ARPM) to enhance features representation and capture both global and local features using different receptive fields to improve the accuracy and robustness of forgery localization. Expensive experiments on 5 public datasets have shown that our proposed model outperforms several the state-of-the-art methods, specially on small region forged image.
\end{abstract}


\end{frontmatter}




\section{Introduction}

In the era of rapid development of information technology, image forgery has become easy. Especially with the help of multimedia tools, anyone can quickly generate an undetectable forged image. Some common modifications such as blurring \cite{blur}, sharpening, flipping, contrast enhancement, etc. do not change the image content and have little effect on its sensitive information. However, forgery with modifing the image content can distort the semantics of the image. Malicious tamperers may use forged images to convey misleading and malicious information to the society through the rapid dissemination of the Internet, especially in the fields of social media, news reporting, and legal evidence, e.g., forged images may be used to misinform the public about certain events, or be used in financial fraud, etc. When malicious forgery is used for images with extremely sensitive content, such as military images, legal evidence, and political news, the malicious impact is immeasurable.

\begin{figure}[!t]
	\centering
	\includegraphics[width=3.5in]{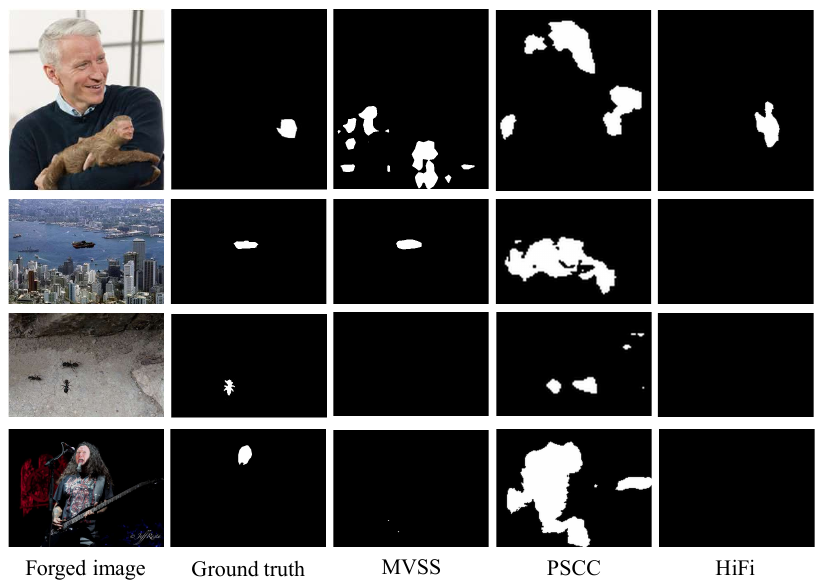}
	\caption{Examples of small forged regions localization. Most  methods subject to false alarms or missed alarms.}
	\label{small}
\end{figure}

There are three common types of image cotent changing forgery: splicing {\cite{Splice}}, copy-move {\cite{CopyMove1}}, removal {\cite{PassiveForgery}}, Splicing refers to copying part of the content of an image and pasting it into another image. Copy-move copies a particular region of the image to a different position on the same image. Removal is to delete a specific object of the image. In order to make the image look natural, the blank area after deleting the object is usually filled by image restoration techniques such as texture synthesis and interpolation. Because copy-move is operated on the same image, the copy-moved elements usually match the surrounding environment in terms of color, lighting, and noise. Therefore, if handled properly, copy-move may be difficult to detect. With the rapid development of image restoration techniques such as Generative Adversarial Networks (GAN) \cite{GAN}, images that have undergone removal operations are increasingly difficult to identify. Images processed by different forged methods will have different forgery features, so more comprehensive forgery features need to be learned to identify the authenticity of images in this case. Moreover, when forged images are spread on the Internet, they are usually subject to post-processing operations such as image compression by the spreading application. This makes it more difficult to detect the forged part of the image. 

To date, numerous studies have focused on IFL based on deep learning techniques such as Convolutional Neural Network (CNN)  and Transformer. Although deep neural network has excellent performance in multimedia understanding and computer vision, they may face some challenges in IFL. For example, the convolutional layer abstracts the image content by reducing the spatial resolution of the feature map through multi-layer convolution, while this approach helps to capture the global information, and the local features are easily lost and feature degradation occurs, especially when dealing with small forged region. On the other hand, although the transformer block is able to capture long-range dependencies, its self-attention mechanism may ignore important local features, thus affecting the accuracy of detection. Fig. {\ref{small}} show some small forged regions localization results of several state-of-the-art IFL methods \cite{MVSS,PSCC,HiFi}. It is easy to find that most of them can not accurately localize the forged regions, even completely fail in some cases.

To cope with these problems, we propose a novel IFL method based guided noise and multi-scale feature aggregation. We first propose a guided noise extractor, in which the guided filter {\cite{GF1}} and Sobel filter {\cite{Sobel}} are dopted to extract different types of forged trace and enhance the edge information in a guied way. Then, the input RGB image and its guided noise are fed into EfficientNetV2 \cite{EffV2} backbone network to learn the RGB and noise feature, respectively. EfficientNetV2 significantly improves the training speed as well as the localization accuracy by optimizing the network structure, using a progressive learning strategy, and dynamically adjusting the regularization method.

To fully integrates the complementary information in the RGB and noise domains, the Feature Aggregation Module (FAM) is designed for aggregating RGB features and noise features using dynamic convolution {\cite{DC}}. Moreover, an Atrous Residual Pyramid Module (ARPM) is presented to expand the receptive field through atrous convolution {\cite{Atrous}} and residual concatenation {\cite{ResNet50}}. ARPM can learn both global and local features and mitigate the feature degradation problem for more comprehensive learning of forgery features. Finally, the processed features are fed into localization modules consisting of attention mechanism {\cite{Attention}}, and generate corresponding masks through spatial and channel attention mechanisms.

In summary, our main contributions are summarized as follows:
\begin{itemize}
	\item[$\bullet$]We propose a novel guided and multi-scale feature aggregated network architecture for forgery image detection and localization. The guided noise features are extracted in a guided way by combining guided filter and Sobel filter, which are enable to learn different types of forged artifacts and edge information. 
	\item[$\bullet$] We propose the Feature Aggregation Module (FAM), which improves feature expressive by using dynamic convolution, enables more efficient aggregation of RGB and noise features.
	\item[$\bullet$] To learn both global and local features, we propose the Atrous Residual Pyramid Module (ARPM), which expands the receptive field using artous convolution and residual concatenation. The different receptive fields enable the network to learn different levels of features of the forged image, thus allowing the model to efficiently localize images with different sizes of forged regions and improving the robustness of it.
	\item[$\bullet$] Extensive experiments on 5 publicly available forgery image datasets shown that our proposed method is superior to the current IFL methods.
\end{itemize}

The remainder of the paper is organized as follows. Section \ref{relate} briefly introduces the related works. Section \ref{method} presents the proposed guided and multi-scale feature aggregated network for IFL. The experimental parameter settings are given in Section \ref{experiment}. Section \ref{result} reports and analyzes the experimental results. Conclusion is made in Section \ref{conclusion}.

\section{Related Work}
\label{relate}
Image forgery localization methods can be broadly divided into traditional methods and deep learning methods.

\begin{table}[t]
	\setlength{\tabcolsep}{1pt}
	\caption{\textcolor{black}{Summary of SOTA methods for image forgery localization. (Eff.V2 denotes EfficientNetV2)}}
	\centering
	\renewcommand{\arraystretch}{1}
	\begin{tabularx}{\columnwidth}{@{}l*{5}{>{\centering\arraybackslash}X}@{}}
		\toprule
		\multirow{2}{*}{Method} & \multirow{2}{*}{Publish} &\multicolumn{2}{c}{Branch} & \multirow{2}{*}{Fusion} & \multirow{2}{*}{Backbone} \\
		\cmidrule(lr){3-4}
		&&RGB & \multicolumn{1}{c}{Noise} &  & \\
		\midrule
		J-LSTM{\cite{JLSTM}}&\small \textit{ICCV2017}&\checkmark&\multicolumn{1}{c}{\small -}&- &LSTM\\
		RGB-N{\cite{RGBNet}}  &\small \textit{CVPR2018}&\checkmark&\multicolumn{1}{c}{\small SRM Filter}&Late&CNN\\
		H-LSTM{\cite{HLSTM}}&\small \textit{TIP2019}&\checkmark&\multicolumn{1}{c}{\small -}&- &LSTM\\
		ManTra{\cite{ManTra}}&\small \textit{CVPR2019}&\checkmark&\multicolumn{1}{c}{\small BayarConv, SRM Filter}&Early&VGG \\
		SPAN{\cite{SPAN}}&\small \textit{ECCV2020}&\checkmark&\multicolumn{1}{c}{\small BayarConv, SRM Filter}&Early&VGG \\
		CAT{\cite{CAT}}&\small \textit{WACV2021}&\checkmark&\multicolumn{1}{c}{\small DCT}&Mid &HR-Net \\
		PSCC{\cite{PSCC}}&\small \textit{TCSVT2022}&\checkmark&\multicolumn{1}{c}{\small -}&-&HR-Net \\
		MVSS{\cite{MVSS}}&\small \textit{TPAMI2023}&\checkmark&\multicolumn{1}{c}{\small BayarConv}&Late&ResNet \\
		HiFi{\cite{HiFi}}&\small \textit{CVPR2023}&\checkmark&\multicolumn{1}{c}{\small Laplacian}&Early&HR-Net \\
		EMT{\cite{EMTNet}}&\small \textit{PR2023}&\checkmark&\multicolumn{1}{c}{\small BayarConv, SRM Filter}&Late&ResNet \\
		EVP{\cite{EVP}}&\small \textit{CVPR2023}&\checkmark&\multicolumn{1}{c}{\small High-frequency Components}&Late&SegFormer \\
		Zhou.{\cite{Edgebased}}&\small \textit{ESWA2024}&\checkmark&\multicolumn{1}{c}{\small Segment-based}&Late&CNN \\
		Ours&-&\checkmark&\multicolumn{1}{c}{\small Guided Filter, Sobel Filter}&Late&Eff.V2\\
		\bottomrule
		\label{summary}
	\end{tabularx}
	\vspace{-1em}
\end{table}

Traditional image processing detection methods mainly rely on manually created or pre-determined features. Lin et al. {\cite{DCT}} proposed to use the statistics of discrete cosine transform coefficients of double compressed JPEG images to distinguish between real and forged regions. However, this method is mainly for JPEG images for forgery detection and may not be applicable for other image formats. Ferrara et al. {\cite{CFA}} and others used Color Filter Array (CFA) to detect inconsistencies in forged regions of an image. This method may be limited if the forgery operation does not leave obvious CFA artifacts or the forged image is post-processed to remove these artifacts. Cozzolino et al. {\cite{Splicebuster}} used local noise introduced by the sensor and its post-processing as a cue for image splicing detection to detect feature inconsistencies or anomalous spliced boundaries between real and forged regions. While it is mainly targeted at splicing images and unapplicable for other forgery types. Vaishnavi et al. \cite{Copy-move_Tradition} proposed a scheme to detect copy-move forgery by means of symmetry based local features ,this method can detects and localizes a plurality of copy-move regions in a single image. Meena et al. proposed a forgery detection technique \cite{Tetrolet} for copy-move images based on the Tetrolet transform, which is used to extract features to localize the copy-move region. Soni et al. proposed a block-based copy-move forgery detection enhancement method \cite{SURF} using hybrid local feature extraction to localize the forged regions of an image by calculating the SURF features of the image and removing outliers.

In recent years, deep learning based methods have become a hot research topic. By training deep neural networks, such as CNN {\cite{CNN}} and GAN {\cite{GAN}}, the models are able to learn high-level features of image forgery, thus improving the accuracy of detection. These methods are not only able to handle more complex and covert image forgery scenarios, adapt to evolving forgery techniques, and outperform traditional methods in terms of performance. As shown in Table \ref {summary}, the current state-of-the-art methods are summarized according to feature types, aggregation methods, backbone and other indicators. Zhou et al. designed a two branches network architecture RGB-N {\cite{RGBNet}} using Faster R-CNN {\cite{FasterRCNN}} and Steganalysis Rich Model (SRM) {\cite{SRM}}.The RGB branch learns image content features to find information such as possible forgery boundaries, and the noise branch aims to discover inconsistencies between real and forged regions. Since it is designed based on Faster R-CNN, the model can only provide bounding boxes and cannot achieve pixel-by-pixel localization. Wu et al. proposed a new scheme ManTra-Net {\cite{ManTra}}, to consider image forgery localization as an anomaly detection task. ManTra-Net categorizes forgery into 385 types, designs Z-score features and uses auxiliary features, such as camera sensors, to distinguish between real and forged regions. Hu et al. improved ManTra-Net and proposed SPAN {\cite{SPAN}}, which constructs the spatial correlation of features through a local self-attentive pyramid model, but the accuracy of SPAN is poor for lower resolution images. Dong et al. proposed MVSS-Net {\cite{MVSS}}, which is a two-branch network use ResNet50 {\cite{ResNet50}} as backbone for both the RGB branch and the noise branch. an additional edge-supervised branch is used in the RGB branch to learn edge artifact information to assist the network in learning forgery features, and BayarConv {\cite{BayarConv}} proposed by Bayar et al. is used in the noise branch to extract the noise. MVSS-Net improves the detection accuracy and prevents false alarms on real images. Liu et al. proposed PSCC-Net {\cite{PSCC}}, which uses HRNet {\cite{HRNet}} as backbone to learn multi-scale image features, and designed SCCM module to learn the correlation between spatial and channel in both spatial and channel dimensions through the attention mechanism {\cite{Attention}}, then generate masks in a progressive form. However, the robustness of PSCC-Net is poor due to the excessive focus on unnecessary features in the SCCM module. Guo et al. proposed HiFi-Net \cite{HiFi}, which uses a coding tree form to localize a total of 13 forgery types including GAN-generated and traditional editing methods, and detects them using a multi-branch feature extractor. HiFi-Net can be considered as an improvement of PSCC-Net. Lin et al. proposed EMT-Net \cite{EMTNet}, which uses ResNet to learn RGB features, Transformer to learn noise features, and Edge Augmentation Module to learn forgery traces. S. Bharathiraja et al. proposed an Anti-Forensic Contrast Enhancement Detector scheme \cite{SVM} that utilizes gamma-corrected second-order derivatives as features and robustly classifies these features on a support vector machine. However, this method can only detect whether the image has been forged or not, and cannot localize the forged region. Considering that the forged region is captured from a different camera than the real region, or from a different image position of the same camera, Cozzolino et al. proposed the noise extractor NoisePrint \cite{NoisePrint}, which learns the imaging features of the camera by self-supervised learning to recognize the inconsistency between the forged region and the real region, and extended it in TruFor \cite{TurFor} by proposing NoisePrint++, which is used in conjunction with RGB images in the two-branch CMX \cite{CMX} architecture. Xia et al. proposed DFMM-Net \cite{DFMM}, which uses a two branches architecture to categorize features with different scales into primary, intermediate, and advanced features. Different aggregation methods are used for different levels of features to utilize the different scales features, which improves the localization performance. Ding et al. proposed AFTLNet \cite{AFTLNet} with adaptive blocks to maximize the forgery trace by iteratively updating the weights using adaptive differential convolution, but this method is only suitable for detecting the removal image. 

There have also been attempts to combine CNNs with LSTMs \cite{LSTM} for IFL. Bappy et al. proposed the J-LSTM \cite{JLSTM} and H-LSTM \cite{HLSTM} models, to capture long-range dependencies in forged images using LSTM, which helps to mine potential forgery traces. However, they are very time consuming, and the detection of the forged region is limited by the size of image patches. Shi et al. proposed PL-GNet \cite{PLGNet}, which unites a two-branch network and LSTM, where noise is extracted using a 7-channels SRM in the noise branch.

\section{Proposed Method}
\label{method}
\subsection{Overview}
\begin{figure*}[htbp]
	\centering
	\includegraphics[width=4.5in]{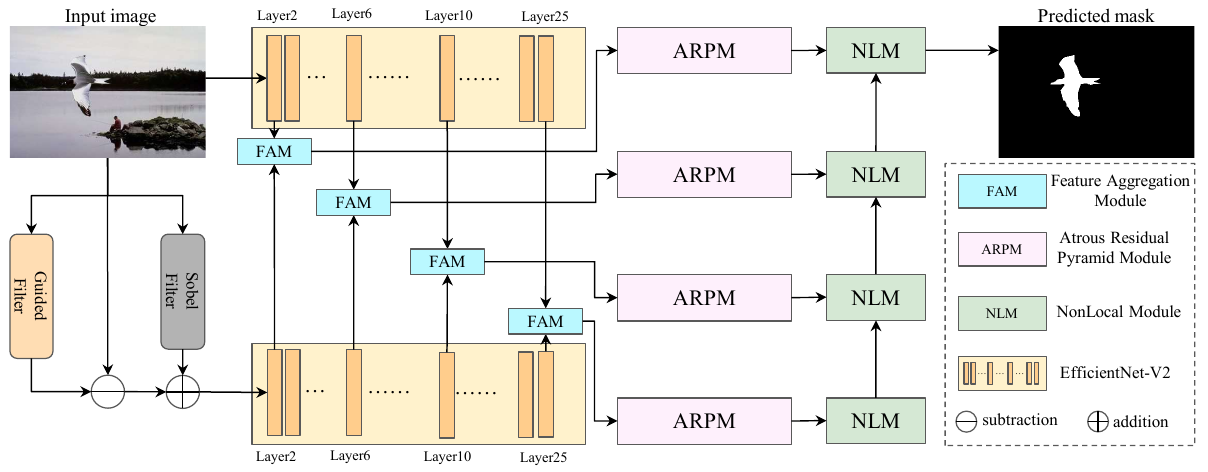}
	\caption{Architecture of the proposed IFL model. It contains a guiede niose extracor, a feature aggregation module, an atrous residual pyramid module and a localization module.}
	\label{fig_2}
\end{figure*}

To improve the localization performance with small forged regions, we propose a novel end-to-end network framework to learn different types of forged artifacts and enhance the feature representation. The architecture of the proposed network for IFL is shown in Fig. \ref {fig_2}.
The input image $I$ is first pre-processed to obtain the guided noise, which can capture different types of forged traces and the edge information. Then, $I$ and its guided noise $I_g$ are fed into the backbone network to learn the RGB and noise features, respectively. In Section \ref{subsec:FAM}, to fully integrate RGB and noise features, the FAM is designed for improving feature expressive by using dynamic convolution. Moreover, the ARPM is presented to learn both global and local features and mitigate the feature degradation problem for more comprehensive learning of forgery features (see Section \ref{subsec:ARPM}). Finally, in the Section \ref{subsec:NLM}, the processed features are fed into localization modules $\text{NLM}$ to obtain the predicted mask. The workflow of the whole network can be expressed by:
\begin{equation}
	M=\operatorname{NLM}(\operatorname{ARPM}(\operatorname {FAM}(\operatorname {B}(I,I_g))))
\end{equation}
where $\text{B}$ denotes the backbone network, $\text{NLM}$ denotes the localization module, and $M$ denotes the predicted mask. In the next sections, each of the above modules will be described.

\subsection{Features extraction}
\subsubsection{RGB features}
In previous works \cite{PSCC,HiFi,CAT,SCFE}, HRNet \cite{HRNet} is widely adopted as the backbone network for IFL. It uses a parallel approach to iteratively fuse multiple scales by concatenating convolutional streams from high resolution to low resolution, so that the learned high-resolution representations are not only semantically strong but also spatially accurate. Although HRNet can capture spatial features ranging from coarse to fine, it is not sufficient to capture more complex or abstract semantic features, moreover, it is resourece intensive and time consuming. In this paper, we employ EfficientNetv2 \cite{EffV2} as the backbone network, which is optimized based on EfficientNet \cite{Eff}, using a simpler and more efficient convolutional structure. Furthermore, we adopt a progressive training method during the training process to improve the performance and efficiency of the model. 

EfficientNetV2 contains features of multiple scales which contain different information, respectively. The shallow network has a smaller receptive field, so the larger scale shallow features contain more detailed information, enabling the network to capture more local features. As the number of convolutions increases, the deep network has a larger receptive field, at which point a pixel point usually contains information about a region, and thus the lower scale deep features contain global features with richer semantics {\cite{DFMM}}. As shown in Fig. \ref {fig_2}, we feed the input image $I$ directly into EfficientNetv2. In order to obtain features at different scales, we select the outputs of four layers, namely, layer 2, layer 6, layer 10, and layer 25, as the outputs of RGB features, respectively, denoted as $\left\{f_{rgb}^1,f_{rgb}^2,f_{rgb}^3,f_{rgb}^4\right\}$. The larger scale feature map contains local information of the input image, such as basic shapes and structures. As the scale of the feature maps decreases, the number of channels gradually increases, these feature maps contain more complex feature information. The smallest scale feature map have the largest number of channels, and contains the global information of the image.

\subsubsection{Noise features}
In the IFL task, the noise in the image also carries forgery traces, which is usefull for localizing forgery \cite{RGBNet,MVSS,HiFi,PLGNet}.
However, for different forgeries, their characteristics are quite different. For example, the splicing forgery copies a region from an image and pastes it onto another image, so that the real region of the spliced image and the forged region have different statistical characteristics of different images. In addition, the copy-move forgery is an internal forgery operation on the same image, the statistical features of the original region and the forged region will be very similar, especially in the presence of post-processing, and it will be more difficult to learn the forged noise information. To cope with this, we propose to extract noise information in a guided way. 

The input forged image $I$ can be considered as a combination of content information $I_c$ and forgery information $I_f${ \cite{SNIS,GF,PRNU}}, as:
\begin{equation}
	I=I_{c}+I_{f}
	\label{GF1}
\end{equation}
According to Eq. \ref{GF1}, if $I_{c}$ is known, the forgery information $I_{f}$ can be easily obtained. Based on this, We first utilize the guided filter to extract the noise information, which is used in image denoising. When the input image is used as the guidance image, it becomes an edge preserving filter that removes noise as well as preserves edges. By performing guided filter, the image content 
$I_{c}$ is preserved and the forgery information $I_{f}$ is eliminatesd. Therefore, the output of the guided filter can be approximated as the image content $I_{c}$ and the forgery information can be given by:
\begin{equation}
	I_{f}=\left| I-\text{Guide}\left( I\right) \right|
	\label{GF3}
\end{equation}
Guided filter is based on a local linear model \cite{GF1}, and subtracting the input image from the output of the guided filter essentially performs a high-pass filtering operation, which can help to highlight high-frequency components of the image, such as texture, forged traces, etc.

In the process of image forgery, any forgery operation will leave edge artifacts and post-processing operations will weaken the edge artifacts and forgery traces, therefore, we use Sobel filter \cite{Sobel} to further extract the edge information. Combining guided filter and Sobel filter enables the network to learn both forgery traces and edge artifacts of the image, which weakens the impact of post-processing operations on the localization performance and enables the network to accurately localize the forged region.

The process of guided noise extractor can be represented as:
\begin{equation}
	I_g=I_{f}+I_{s}=I_{f}+\text{Sobel}\left( I\right)
\end{equation}
where $I_g$ denotes the guided noise, and $I_{s}$ is the edge feature obtain by Sobel filter operation Soble($\cdot$).

Then, similar to RGB features, the guided noise $I_{g}$ is sent to EfficientNetv2 to learn the noise features, and the final output of the noise features with four different scales is recorded as $\left\{f_{n}^1,f_{n}^2,f_{n}^3,f_{n}^4\right\}$, they also have different scales, where large scale shallow feature maps are used to capture information of the input image such as edges, texture, etc., and as the scale of the feature maps decreases, small scale feature maps are used to capture global features in the image as well as inconsistencies in the noise features.

\subsection{Feature Aggregation Module}
\label{subsec:FAM}
\begin{figure}[!h]
	\centering
	\includegraphics[width=3.5in]{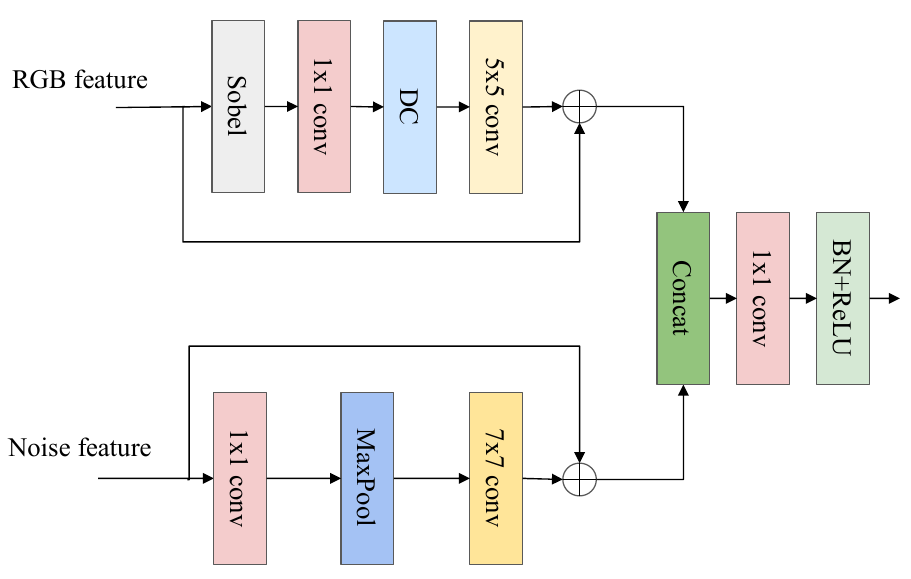}
	\caption{The architecture of the Feature Aggregation Module}
	\label{FFM}
\end{figure}

In two branches architecture network, feature aggregation is very important, which can improve the sensitivity of
the features to different types of forgeries, and enhance the localization ability and robustness to  forged regions with different sizes. Most of the methods in previous work are directly concatenate RGB and noise features in the channel dimension, or add the two features together, which makes the correlation between the RGB and noise features not fully utilized. When there is a large difference between the features, some important forgery information may be ignored, resulting in the network's difficulty in learning effective forged features.
To mine the latent complementary relationships between the RGB features and the noise features, we design a features aggregation module, as shown in Fig. \ref{FFM}.

For RGB features, we first use a Sobel filter to enhance the edge information extracted from the backbone network, and reduce the number of channels through a $1\times1$ convolution. Then it is fed into a $3\times3$ Dynamic Convolution (DC){ \cite{DC}}. DC improves the representation of features by fusing multiple convolution kernels to enhance the model performance without increasing the depth and width of the lifting network, and it can also dynamically generate convolution kernels based on the input data, allowing the convolution operation to adapt itself to different inputs. By dynamically adapting the convolution kernel, the network is better able to handle the different forgery traces produced by different types of forgeries, improving localization performance. Moreover, DC is able to capture multi-scale and multi-type features at the same level. For complex forgery traces present in different types of forgeries, dynamic convolution can flexibly adjust the convolution kernel to effectively learn the forgery traces. Then after a $5\times5$ convolution, it expands the feature receptive field to capture a larger range of feature patterns. Finally, the processed feature are summed up with the input feature by residual concatenation, which effectively combines the original features with the edge features to learn different forgery traces and enhance the expressive ability of RGB features. The above process is expressed by the equation as:
\begin{equation}
	\hat{f}_{rgb}^i=\operatorname{C}_{5 \times 5}(\operatorname{C_{D}}(\operatorname {C}_{1 \times 1}(\operatorname {Sobel}(f_{rgb}^i)))) + f_{rgb}^i
\end{equation}
where $\hat{f}_{rgb}^i$ denotes the enhanced RGB features in the feature aggregation module, $\mathrm{C_{k \times k}}$ denotes the $\mathrm{k \times k}$ convolution layer, $\mathrm{C_D}$ denotes the DC, and $f_{rgb}^i$ denotes the RGB features of the input FAMs, $i \in \{1, 2, 3, 4\}$, respectively.

Similar to the processing of RGB features, we first use a $1\times1$ convolution to reduce the number of channels of features. Then, a Max Pooling is employed to squeeze the feature dimensions and retain important features. Besides, the features are globally sensed by a $7\times7$ convolution to capture a wider range of contextual information. Next, the processed features are summed up with the input features by residual concatenation. The above process can be written as:
\begin{equation}
	\hat{f}_{n}^i=\operatorname{C}_{7 \times 7}(\operatorname{Max}(\operatorname {C}_{1 \times 1}(f_{N}^i))) + f_{n}^i
\end{equation}
where $\hat{f}_{n}^i$ denotes the enhanced noise features, and $\mathrm{Max}$ is the Max Pooling layer.

Finally, the enhanced RGB features and noise features are concatenated in the channel dimension and then followed by a $1\times1$ convolution, a batch normalization layer, and a ReLU activation function. The aggregated feature is expressed by:
\begin{equation}
	{f}_{Agg}^i=\operatorname{ReLU}(\operatorname{BN}(\operatorname {C}_{1 \times 1}(\hat{f}_{rgb}^i : \hat{f}_{N}^i))
\end{equation}

\subsection{Atrous Residual Pyramid Module}
\label{subsec:ARPM}
For forged images, the size of forged regions varies, small forged regions tend to have low resolution due to the small area they occupy. The traces of forgery contained in small forged regions after multiple convolutions are easily ignored and blended with the background, making it difficult to accurately localize them, which results in feature degradation problems. Using a feature pyramid structure{ \cite{FPN}} allows features to be extracted at different levels and effectively capturing feature information from small forged regions. Therefore, we propose ARPM to capture global and local features of forged images and to enhance the representation of the features to improve the localization of small forged regions.
\begin{figure}[!h]
	\centering
	\includegraphics[width=3.5in]{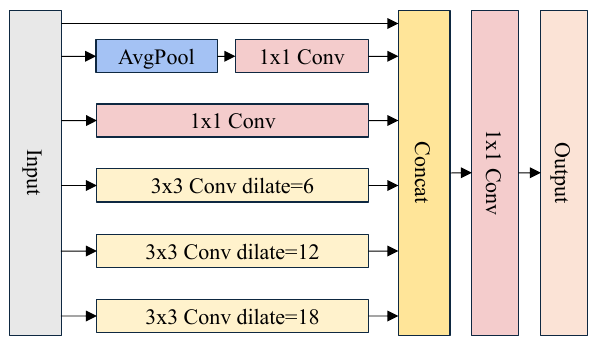}
	\caption{The architecture of the Atrous Residual Pyramid Module}
	\label{ARPM}
\end{figure}

The structure of ARPM is shown in Fig. \ref{ARPM}., where the input features are fed into an global average pooling layer, a 1$\times$1 convolutional layer, to learn a more abstract and high-level feature representation, enhancing the feature representation, which can be given by:
\begin{equation}
	f_{avg} = \mathrm{C_{1 \times 1}}(\mathrm{GAP}(f_{Agg}))
\end{equation}
where $f_{Agg}$ denotes the feature after FAM aggregation, $\mathrm{GAP}$ denotes the global average pooling layer. At the same time, the input features are also fed into a 1$\times$1 convolutional layer, three 3$\times$3 atrous convolutions with dilation of 6, 12 and 18, respectively, which can be given by:
\begin{equation}
	f_{1 \times 1} = \mathrm{C_{1 \times 1}}(f_{Agg})
\end{equation}
\begin{equation}
	f^{i}_{d} = \mathrm{C}^{d=i}_{3 \times 3}(f_{Agg}),  i = \{6, 12, 18\}
\end{equation}
where $\mathrm{C}^{d=i}_{3 \times 3}$ denotes the convolutional layer with convolutional kernel 3 and dilation rate $i$. Atrous convolution{ \cite{Atrous}} captures a wider range of contextual information by increasing the receptive field of the convolution kernel, which can keep the resolution of the feature map unchanged while increasing the receptive field compared to traditional convolution operations. We use multiple convolutional layers with different dilation rates to learn features at different levels of the image, including low-level features $\left(\text{edges and textures}\right)$ as well as high-level features $\left(\text{semantic information}\right)$, through different receptive fields. Finally, the input features and the outputs of the above five layers are concatenated in the channel dimension and using a 1$\times$1 convolutional layer to reducing the number of channels. ARPM can be expressed by the equation as:
\begin{equation}
	X=\mathrm{C_{1 \times 1}}\left(f_{Agg}:f_{avg}:f_{1 \times 1}:f^{6}_{d}:f^{12}_{d}:f^{18}_{d}\right)
\end{equation}

\subsection{Localization Module}
\label{subsec:NLM}
In the above work, The IFL Network we proposed learned the forged features through the backbone network, FAM and ARPM, which contain rich global and local features and have edge artifact features that contain traces of forgery, mitigates the feature degradation problem. These features are then used to locate the forged area via the localization module. 

In IFL tasks, the attention mechanism \cite{Attention} is generally used to localize the forged region based on the processed features, e.g., HiFi-Net \cite{HiFi} and PSCC-Net \cite{PSCC} both use NonLocal Module (NLM) \cite{NonLocal} or the variant of NLM for localization. Attention mechanism can help the network to better learn the feature representation and improve the detection accuracy by emphasizing important features and suppressing irrelevant information \cite{Attention}, attention can be classified into spatial attention \cite{NonLocal} and channel attention \cite{SENet}. We use the variant of the NLM proposed in PSCC-Net \cite{PSCC}, the Spatial-Channel Correlation Module (SCCM), as the localization module of the network, which fuses spatial and channel attention. Fusing spatial and channel attention can effectively improve localization accuracy.

In SCCM, the input feature $X \in \mathbb{R}^{C \times H \times W}$ is reshaped into $X' \in \mathbb{R}^{\frac{HW}{r^2} \times Cr^2}$. This operation greatly reduces the amount of spatially relevant computation while preserving all feature information. We then use three \(1 \times 1\) convolutions \(g\), \(\theta\), \(\varphi\) to transform the input features $X^{'}$ into $X_{g} = g(X^{'})$, $X_{\theta} = \theta(X^{'})$, $X_{\varphi} = \varphi(X^{'})$, and compute the spatial attention matrix $m_s$ and the channel attention matrix $m_c$. 
Finally, the spatial correlation attention $A_s$ and the channel correlation attention $A_c$ can be given by:
\begin{equation}
	A_{s}=m_{s} \times X_{g}=\operatorname{softmax}\left(X_{\theta} \times X_{\varphi}^{T}\right) \times X_{g}
\end{equation}
\begin{equation}
	A_{c}=X_{g} \times m_{c}=X_{g} \times \operatorname{softmax}\left(X_{\theta}^{T} \times X_{\theta}\right)
\end{equation}
where $A_s$ is the feature generated from spatial attention and $A_c$ is the feature generated from channel attention. Finally we reshape the size of $A_s$ and $A_c$ to $C \times H \times W$. The feature representation is enhanced by using two $1\times1$ convolutional layers $k_s$ and $k_c$, and the spatial and channel attention is summed with the input features and residual connections to obtain the feature $F$ used to generate the prediction mask:
\begin{equation}
	F=X+s \cdot k_{s}\left(A_{s}\right)+c \cdot k_{c}\left(A_{c}\right)
\end{equation}
where $s$ and $c$ denote the weights and $s+c=1$.

The final one-channel binary mask $M$ is finally generated by a $3\times3$ convolution and sigmoid function:
\begin{equation}
	M=\operatorname{Sigmoid}(\operatorname{C}_{3 \times 3}(\operatorname {ReLU}(\operatorname{C}_{3 \times 3}(F)))
\end{equation}
where $\mathrm{Sigmoid}$ and $\mathrm{ReLU}$ denote the activation functions

The network contains a total of four localization modules, where each localization module takes the output of the previous localization module as a priori information and gradually generates the exact mask in progressive mechanism. We denote the four different scales of prediction masks output from the four localization modules as $\left\{M_{1},M_{2},M_{3},M_{4}\right\}$ respectively, and output $M_{4}$ is taken as the final predicted mask.

\subsection{Loss Function}
In the training process, the image is divided into forged and real regions according to the ground-truth, in which 1 and 0 denote the pristine pixel and forged pixel respectively. Therefore, the forgery localization can be regarded as a binary classification task with the goal of distinguishing between forged and pristine pixels. In the forgery image, the forged region occupies only a small portion of the image, which means that the labels are highly unbalanced at the pixel level. To address this problem, the binary cross-entropy loss $\left( \mathrm{BCE\_Loss} \right)$ is adopted in the four localization modules. $\mathrm{BCE\_Loss}$ facilitates the model's learning to distinguish between these unbalanced classes by calculating the loss for each pixel's prediction individually. Finally, the total loss is denoted as:
\begin{equation}
	L = \sum_{i=1}^{4} \text{BCE\_Loss}(M_i, G_i)
\end{equation}
where $M_i$, $G_i$ denote the corresponding predict mask and ground-truth, respectively, $i \in \{1, 2, 3, 4\}$. In addition, we consider the losses of the four localization modules are of equal importance, and therefore give them the same weight.

\section{EXPERIMENTS}
\label{experiment}
In this section, we conduct extensive experiments on five publicly available standard datasets to compare with current state-of-the-art methods and evaluate the performance of our proposed model. In addition, we also perform robustness and ablation experiments. 
\subsection{Datasets}
\begin{table}[h]
	\caption{\textcolor{black}{5 Test datasets configurations and fine-tuned segmentation}}
	\centering
	\renewcommand{\arraystretch}{1}
	\begin{tabularx}{\columnwidth}{>{\centering\arraybackslash}X >{\centering\arraybackslash}X >{\centering\arraybackslash}X >{\centering\arraybackslash}X}
		\toprule
		\multirow{2}{*}{Dataset} & \multirow{2}{*}{Pre-Train} & \multicolumn{2}{c}{Fine-tune} \\
		\cmidrule(lr){3-4}
		& & Train & Test \\
		\midrule
		Columbia&180&-&- \\
		Coverage&100&75&25 \\
		CASIA&6044&5123&921 \\
		NIST16&564&-&- \\
		IMD20&2010&-&- \\
		\bottomrule
	\end{tabularx}
	\label{dataset}
\end{table}

\subsubsection{Train Data} 
To facilitate the evaluation of performance, we use the same publicly available dataset as \cite{PSCC} for training. The training dataset contains a total of four types of images: 1) real, 2) copy-move, 3) splicing,  4) removal. Among them, real, spliced and removed images are generated by the Microsoft COCO dataset \cite{MSCOCO}. Images of copy-move type are obtained from \cite{CopyMove}. This dataset contains 82k real images, 100k copy-move images, 116k splicing images and 78k remove images totaling 376k images.

\subsubsection{Test Data}
In order to compare with the current state-of-the-art methods, we used five publicly available datasets, which are Columbia \cite{Columbia}, Coverage  \cite{Coverage}, CASIA \cite{CASIA}, NIST16 \cite{NIST16}, and IMD20 \cite{IMD20}, as our test sets. Table \ref{dataset} shows the number of images in these five datasets, as well as the fine-tuned divisions. The five datasets are described below.
\begin{itemize}
	\item[$\bullet$] \textit{Columbia}: Columbia contains 180 forged images with dimensions of 757$ \times $568 pixels and the corresponding masks are provided. the forgery type of all images is splicing. for the pre-trained model, all images are used for testing.
	\item[$\bullet$] \textit{Coverage}: Coverage contains 100 forged images with an average size of 400$ \times $486 pixels, and the corresponding original images and masks are provided. All the images are also rotated, scaled, and other six different post-processing processes. For the pre-trained model, all images were used for testing and for the fine-tuned model, we divided the dataset according to the scale, where 75 images are used for training and 25 images are used for testing.
	\item[$\bullet$] \textit{CASIA}: The CASIA dataset contains two versions: CASIA V1.0 and V2.0. CASIA V1.0 contains 921 forged images, V2.0 contains 5123 forged images, most of the images are of size 384$ \times $256 pixels and the corresponding mask is provided.The forgery types are splicing and copy-move. For the pre-trained model, all images were used for testing and for the fine-tuned model, V2.0 is used for training and CASIA V1.0 is used for testing.
	\item[$\bullet$] \textit{NIST16}: The NIST16 dataset contains 564 forged images with an average size of 3460$ \times $2616 pixels, the corresponding masks are provided. The forgery types are copy-move, remove, and splicing. For the pre-trained models, all the images are used for testing.
	\item[$\bullet$] \textit{IMD20}: The IMD20 dataset consists of 2,010 real forged images collected from the internet with an average size of 1056$ \times $848 pixels. The forgery types are copy-move, remove and splicing. For the pre-trained model, all images are used for testing.
\end{itemize}

\subsection{Experimental Setup}
Our model is implemented by the PyTorch framework and trained using NVIDIA Quadro V100 GPU with 32 GB of memory. During the training process, we use Adam \cite{Adam} as optimizer, and reshape all inputs to 256$ \times $256 pixels. In order to make full use of the GPU device, the batchsize is set to 10, the learning rate is 2e-4, and it is halved for every 5 epochs of training, for a total of 25 epochs.

\subsection{Evaluation Metrics}
To quantify the localization performance, based on previous work \cite{HiFi,MVSS,EITLNet}, we use three metrics, AUC, F1 score and IoU, for performance evaluation. AUC denotes the area under the ROC curve, the F1 score is the harmonious average of Precision and Recall, and IoU denotes the ratio of the intersection of the prediction mask and ground-truth area to the area of the concatenated set.AUC, F1 score and IoU all range between 0 and 1, and the closer to 1, the better the performance of the model.

\section{Experiments Results}
\label{result}
This section compares our proposed model with the current state-of-the-art models in terms of localization performance, robustness, etc. In Section \ref{subsec:loc}, we compare the localization performance of our model with the current state-of-the-art IFL methods for all three criteria of AUC, F1 score, and IoU for both pre-trained and fine-tune models. In Section \ref{subsec:small}, we select the dataset of small forged regions to demonstrate the superior performance of our model in localizing small forged regions. In Section \ref{subsec:rb}, we apply various post-processing operations to the images and perform localization tests to compare the robust performance of our model with other methods. In Section \ref{subsec:ab}, we conduct ablation experiments to validate the effectiveness of each of the proposed modules. Finally, in Section \ref{subsec:vis}, we show the results of the localization visualization of our model with other methods on forged regions of different sizes.

\begin{table}[h]
	\centering
	\setlength{\tabcolsep}{5pt}
	\caption{Performance comparison of different forgery localization methods on different datasets (AUC and F1). The best results of per test sets are highlighted in \textbf{bold} and the second-best is \uline{underlined}.  ``-'' indicates not applicable.}
	\renewcommand{\arraystretch}{1.2}
	\begin{tabularx}{\columnwidth}{@{}l*{12}{>{\centering\arraybackslash}X}@{}}
		\toprule
		\multicolumn{1}{c}{\multirow{2}{*}{Methods}} & \multicolumn{2}{c}{Columbia} & \multicolumn{2}{c}{Coverage} & \multicolumn{2}{c}{CASIA} & \multicolumn{2}{c}{NIST16} & \multicolumn{2}{c}{IMD20} & \multicolumn{2}{c}{Average}  \\
		\cmidrule(lr){2-3} \cmidrule(lr){4-5} \cmidrule(lr){6-7} \cmidrule(lr){8-9} \cmidrule(lr){10-11} \cmidrule(lr){12-13} 
		& \small AUC &\small F1 &\small AUC &\small F1 &\small AUC &\small F1 &\small AUC &\small F1 &\small AUC &\small F1 &\small AUC &\small F1  \\
		\midrule
		\small ManTra\cite{ManTra}&\small 82.4 &\small 35.7 &\small 81.9 &\small 27.5 &\small 81.7 &\small 13.0 &\small 79.5 &\small 8.8 &\small 74.8 &\small 18.3 &\small 80.1 &\small  20.7 \\
		\small SPAN\cite{SPAN}&\small 93.6 &\small 81.5 &\small \uline{92.2} &\small 53.5 &\small 79.7 &\small 33.6 &\small 84.0 &\small 29.0 &\small 75.0 &\small 14.5 &\small 84.9 &\small 42.4  \\
		\small PSCC\cite{PSCC}&\small \uline{98.2} &\small \textbf{93.5} &\small 84.7 &\small 49.8 &\small 82.9 &\small 36.3 &\small \uline{85.5} &\small \uline{35.7} &\small \uline{80.6} &\small 19.7 &\small 86.4 &\small 47.0  \\
		\small MVSS\cite{MVSS}&\small 80.9 &\small 66.1 &\small 75.4 &\small 48.3 &\small 80.4 &\small 61.6 &\small 66.0 &\small 29.1 &\small 64.3 &\small 26.0 &\small 79.4 &\small 46.2  \\
		\small EMT\cite{EMTNet}&\small 83.2 &\small 56.1 &\small 81.2 &\small 35.3 &\small 85.6 &\small 45.9 &\small - &\small - &\small - &\small - &\small 83.3 &\small 45.8 \\
		\small HiFi\cite{HiFi}&\small \textbf{98.4} &\small \uline{90.6} &\small \textbf{92.5} &\small \uline{67.6} &\small \uline{86.6} &\small \uline{63.8} &\small 81.1 &\small 28.5 &\small 76.0 &\small \uline{53.2} &\small \uline{87.6} &\small \uline{60.7}  \\
		\small EVP\cite{EVP}&\small 79.1 &\small 27.7 &\small 71.6 &\small 56.9 &\small 86.2 &\small 11.4 &\small 77.5 &\small 21.0 &\small \textbf{81.1} &\small 23.3 &\small 79.1 &\small 28.1  \\
		\small Zhou.\cite{Edgebased}&\small 89.7 &\small 47.0 &\small 72.9 &\small 26.2 &\small 83.6 &\small 33.8 &\small - &\small - &\small 64.2 &\small 24.2 &\small 77.6 &\small 32.8  \\
		\midrule
		\small Ours &\small 93.9 &\small 84.5 &\small 91.4 &\small \textbf{75.7}  &\small \textbf{89.4} &\small \textbf{66.5} &\small \textbf{86.9} &\small \textbf{47.8} &\small 80.2 &\small \textbf{58.9} &\small \textbf{88.4} &\small \textbf{66.7} \\
		\bottomrule
	\end{tabularx}
	\label{AUC_F1_Result}
\end{table}

\subsection{Comparisons on Localization}
\label{subsec:loc}
Comparative IFL methods include ManTra-Net \cite{ManTra}, SPAN \cite{SPAN}, PSCC-Net \cite{PSCC}, MVSS-Net \cite{MVSS}, EMT-Net \cite{EMTNet}, HiFi-Net \cite{HiFi}, EVP \cite{EVP} and the IFL method proposed by Zhou et al. \cite{Edgebased}.All model results are taken from the original paper or run the publicly available source code. 

Table \ref{AUC_F1_Result} reports the F1 score, AUC metrics and average metrics of the pre-trained models on the five datasets. As can be seen from the table, our model achieves the best performance on four datasets, Coverage, CASIA, NIST16, and IMD20, with improvements of 8.1\%, 2.7\%, 19.3\%, and 5.7\%, respectively, compared to HiFi-Net, We attribute the huge improvement on NIST16 to the fact that in the NIST16 dataset, most of the images have small forged regions, and the average forged region area of the NIST16 dataset is the smallest in the five publicly available datasets we use, which is only 7.45\%. We leverage the backbone network to separately extract RGB and noise features from the input image, and efficiently aggregate these features using the FAM. Additionally, the ARPM is employed to capture global features, enabling the network to effectively detect and localize small forged regions, leading to significant improvements in localization performance on the NIST16 dataset.
For the AUC metrics, our model outperforms others, achieving the best results on the CASIA dataset with a 2.8\% improvement over HiFi-Net. Similarly, on the NIST16 dataset, it surpasses PSCC-Net by 1.4\%, and overall, it delivers the highest average performance across all five datasets.

Table \ref{IoU_Result} reports the IoU metrics and average metrics of the pre-trained models on the five datasets. It can be seen from the table that our model achieves the best performance on the Coverage and NIST16 datasets, with an improvement of 10.5\% and 0.7\% compared to the second best model, respectively, while we achieve the second best performance on Columbia. In addition, we also achieved the best performance on the average of the five datasets. 

\begin{table}[h]
	\setlength{\tabcolsep}{10pt}
	\caption{\textcolor{black}{Comparison of IoU results of the pre-trained model, the best performance is shown in \textbf{bold}, the second-best is \uline{underlined}.}}
	\centering
	\renewcommand{\arraystretch}{1}
	\begin{tabularx}{\columnwidth}{@{}l*{6}{>{\centering\arraybackslash}X}@{}}
		\toprule
		Method    & Columbia        & Coverage      & CASIA        & NIST16        & IMD20         & Average \\
		\midrule
		ManTra \cite{ManTra}& 25.8            & 18.6          & 8.0          & 5.4           & 12.4          & 14.0\\
		SPAN \cite{SPAN}      & 39.0            & 10.5          & 11.2         & 15.6          & 10.0          & 17.3 \\
		PSCC \cite{PSCC}  & 36.0            & 13.0          & 23.2         & 10.8          & 12.0          & 19.0 \\
		MVSS \cite{MVSS}  & 57.3            & \uline{38.4}  & \uline{39.7} & 23.9          & \textbf{19.2}  & \uline{35.7} \\
		HiFi \cite{HiFi}  & \textbf{77.2}   & 30.1          & 26.3         & \uline{24.2}  & 8.0           & 33.2 \\
		EVP \cite{EVP}       & 21.3            & 8.3           & \textbf{42.1}& 16.0          & \uline{18.3}         & 21.2 \\
		\midrule
		Ours      & \uline{67.4}   &\textbf{48.9}  & 30.8         &\textbf{24.9}  &17.0  & \textbf{37.8}\\
		\bottomrule
	\end{tabularx}
	\label{IoU_Result}
\end{table}

It is worth the reader's attention that on the NIST16 dataset, our model improves tremendously on the F1, but improves less than the F1 in terms of AUC and IoU. We can easily explain this phenomenon: the NIST16 dataset has a small area of forged regions, the distribution of real and forged regions is not balanced, and the F1 emphasizes the precision and recall of forged regions localization. Our model performs well on small forged regions, so the F1 is significantly improved. The AUC and IoU are global metrics that consider the performance of the network on all samples, so the increase will be lower than the F1.	

For fine-tuning, we initialized the model using the pre-trained model weights and fine-tuned the model on the Coverage and CASIA datasets according to the division ratios reported in Table \ref{dataset}, with all the training procedures for fine-tuning being the same as for the pre-training strategy. 

\begin{table}[htp]
	\setlength{\tabcolsep}{10pt}
	\centering
	\caption{AUC/ F1 score results of the fine-tune model. The best performance is shown in \textbf{bold}, the second-best is \uline{underlined}.}
	\begin{tabularx}{\columnwidth}{@{}l*{6}{>{\centering\arraybackslash}X}@{}}
		\toprule
		\multicolumn{1}{c}{\multirow{2}{*}{Methods}} & \multicolumn{2}{c}{Coverage} & \multicolumn{2}{c}{CASIA} & \multicolumn{2}{c}{Average}   \\
		\cmidrule(lr){2-3} \cmidrule(lr){4-5} \cmidrule(lr){6-7} 
		& AUC & F1 & AUC & F1 & AUC & F1   \\
		\midrule
		SPAN {\cite{SPAN}}& 93.7 & 55.8 & 83.8 & 40.8 & 88.8 & 48.3   \\
		PSCC {\cite{PSCC}}& 94.1 & 72.3 & 87.5 & 55.4 & 90.8 & 63.9 \\
		ObjectFormer {\cite{ObjectFormer}}& 95.7 & 75.8 & 88.8 & 57.9 & 91.9 & 66.9  \\
		HiFi {\cite{HiFi}}& \uline{96.1} & \uline{80.7} & \uline{88.5} & \uline{61.6} & \uline{92.3} & \uline{70.7} \\
		\midrule
		Ours & \textbf{97.4} & \textbf{87.3} & \textbf{89.8} & \textbf{69.2}  & \textbf{93.6} & \textbf{78.3} \\
		\bottomrule
	\end{tabularx}
	\label{AUC_Result_f}
\end{table}

Table \ref{AUC_Result_f} reports the AUC and F1 score performance of the fine-tuned model on the Coverage and CASIA datasets. Our model achieves the best performance on both Coverage and CASIA datasets, improving the AUC and F1 scores by 1.3\% and 7.2\% for the Coverage dataset and 1.3\% and 7.6\% for the CASIA dataset, respectively, compared to HiFi-Net. In addition we achieved the best performance on the average of both AUC and F1 scores.	

The enhancement in localization of our proposed model can be explained by the following reasons. First, PSCC \cite{PSCC} only focuses on the RGB color space of the image and does not learn the noise features, which makes many features related to forgery traces ignored, resulting in poor localization. Second, the methods proposed by ManTra \cite{ManTra}, SPAN \cite{SPAN}, MVSS \cite{MVSS}, EMT \cite{EMTNet}, HiFi \cite{HiFi}, EVP \cite{EVP}, and the method proposed by Zhou et al. \cite{Edgebased} all add additional branches to learn the noise features. However, in the process of fusing the RGB and noise features, these methods are either directly adding the features of the two branches or simply concatenate them in the channel dimension, which prevents the features from being well aggregate. In the process of cross-domain aggregation, incomplete feature aggregation may lead to the hiding of some local features, thus losing certain information related to the forgery traces, which is not conducive to localization of the forged region, as shown in Fig. \ref{visual}. 

Accordingly, we propose FAM for cross-domain feature aggregation in networks. For copy-move type of forged images, the forgery part comes from the image itself, in this case the forgery traces only exist in the edge information, we use Sobel filter to highlight the edge texture, which makes the edge information more prominent in the aggregated features. We also use Dynamic Convolution to process the RGB features, which improves the expressive power of the features by virtue of the Dynamic Convolution's ability to adaptively adjust the properties of the convolution kernel to learn more recognizable features. For noise information, We use a convolutional layer with a convolutional kernel size of 7 to extend the receptive field and learn global noise features. With the above aggregation strategy, we effectively aggregate RGB features and noise features. Next, we use the proposed ARPM to learn different levels of features with different sizes of sensory fields and aggregate them so that the features contain both global and local forgery information, which helps to improve the localization results.

\subsection{Small Forged Regions Performance}
\label{subsec:small}
\begin{table}[h]
	\setlength{\tabcolsep}{1.5pt}
	\caption{\textcolor{black}{AUC/ F1 score results of the small forged regions Localization. The best performance is shown in \textbf{bold}, the second-best is \uline{underlined}.}}
	\centering
	\renewcommand{\arraystretch}{1}
	\begin{tabularx}{\columnwidth}{@{}l*{3}{>{\centering\arraybackslash}X}@{}}
		\toprule
		Method &  Region \(<1\%\)  & Region \(<5\%\)  & Region \(<10\%\) \\
		\midrule
		MVSS {\cite{MVSS}}      & 64.8/38.9 & 75.8/50.9 & 77.8/54.9  \\
		PSCC {\cite{PSCC}}  & \uline{82.9}/52.9 & \uline{83.6}/\uline{55.2} & \uline{83.8}/\uline{56.5}  \\
		HiFi {\cite{HiFi}}   & 81.2/\uline{53.2} & 81.7/54.3 & 81.8/55.2  \\
		\midrule
		Ours       & \textbf{87.1}/\textbf{55.8} & \textbf{90.3}/\textbf{64.1} & \textbf{90.9}/\textbf{66.7} \\
		\bottomrule
	\end{tabularx}
	\label{Small_Result}
\end{table}

Our proposed FAM as well as ARPM have good performance in large forged region localization. In order to verify the effectiveness of our model on small forged areas, we select all forged images in five test datasets with forged areas less than \textbf{1\%}, \textbf{5\%} and \textbf{10\%} in size for testing, respectively. 1157, 3428 and 4667 forged images were selected according to the three small area criteria respectively. Table \ref{Small_Result} reports our test results. Due to the limitations of the dataset, we chose to compare only those methods for which the source code is publicly available. From the table, it is easy to see that thanks to the two modules FAM and ARPM for better learning of features and processing of features through Dynamic Convolution and Atrous Convolution, our AUC and F1 scores are higher than those of MVSS-Net {\cite{MVSS}}, PSCC-Net {\cite{PSCC}}, and HiFi-Net {\cite{HiFi}} for all three small area criteria, on test images with less than 1\% of forged region, we improved 22.3\%, 4.2\% and 5.9\% in terms of AUC compared to them, respectively, it proves the validity of our proposed FAM and ARPM. 

\subsection{Robustness Analysis}
\label{subsec:rb}
\begin{table}[h]
	\setlength{\tabcolsep}{1.5pt}
	\caption{\textcolor{black}{Localization performance (AUC) on Columbia dataset under various distortions.}}
	\centering
	\renewcommand{\arraystretch}{1}
	\begin{tabularx}{\columnwidth}{@{}l*{5}{>{\centering\arraybackslash}X}@{}}
		\toprule
		Distortion  & MVSS {\cite{MVSS}} &SPAN {\cite{SPAN}} & PSCC {\cite{PSCC}} &  HiFi {\cite{HiFi}} & Ours \\
		\midrule
		w/o Dis.   & 80.9& 93.6 & \uline{98.2}  & \textbf{98.4} & 93.8 \\
		\midrule
		Resize(0.78$\times$)   & 62.8 & 85.5 & \uline{93.4}  & 91.4 & \textbf{93.5}(0.3$\downarrow$) \\
		Resize(0.25$\times$)  & 50.1 & 78.4 & 69.0  & \uline{79.8} & \textbf{87.8}(6.0$\downarrow$) \\
		\midrule
		Blur(k=3)   & 69.6 & 78.9& 84.2  & \uline{84.5} & \textbf{87.1}(6.7$\downarrow$) \\
		Blur(k=15)  & 52.6 & 66.7& 73.2  & \uline{78.6} & \textbf{85.9}(7.9$\downarrow$) \\
		\midrule 
		Noise($\sigma$=3)   & 58.1 & 75.1& 82.6  & \uline{88.2} & \textbf{88.9}(4.9$\downarrow$) \\
		Noise($\sigma$=15)   & 56.0 & 60.1& 74.4  & \uline{82.9} & \textbf{87.1}(6.7$\downarrow$) \\
		\midrule
		Comp.(q=100)  & 56.5  & 82.1& \textbf{97.8} & \uline{96.2} & 92.3(1.5$\downarrow$) \\
		Comp.(q=50)  & 54.8 & 73.1& 89.1  & \uline{89.4} & \textbf{89.6}(4.2$\downarrow$) \\
		\bottomrule
	\end{tabularx}
	\label{RB_pscc1}
\end{table}	
In real-life scenarios, during the distribution of images, social network media usually perform a series of post-processing operations on the images, such as Resize, Gaussian noise, Gaussian blur and JPEG compression. These are non-malicious operations, but they are common in real life scenarios. Typically tamperers also perform post-processing operations on the forged images. These post-processing operations blur the forgery traces and make it difficult for the model to localize the forged regions. Therefore robustness is an important performance evaluation metric when the model is applied in real-life scenarios.

We performed 4 different post-processing operations on the images: \textbf{Resize}, \textbf{Gaussian Blur}, \textbf{Gaussian Noise}, and \textbf{JPEG Compression}, and each of them has two different parameters, and each post-processing operation and the corresponding parameter settings are reported in Table \ref{RB_pscc1}, in total, eight post-processing operations are performed, and the model's robust performance is evaluated on the Columbia dataset.

Table \ref{RB_pscc1} reports the robustness performance of our model on the Columbia dataset. It can be seen that our model maintains the highest performance in all post-processing operations except for JPEG compression with a quality factor of 100, e.g., in the test image after Gaussian blurring with k = 15, the AUC of our method is 7.3\% and 12.7\% higher than that of HiFi and PSCC, respectively; and in the test image after Gaussian noise processing with \(\sigma\) = 15, the AUC of our method is 4.2\% and 12.7\% higher than HiFi and PSCC, respectively, which indicates the good robustness of our model. Although PSCC-Net {\cite{PSCC}} and HiFi-Net {\cite{HiFi}} outperform us without post-processing operations, we outperform them after post-processing operations, this is because they overly focus on the global information of the image, which causes the model to focus on irrelevant features. Post-processing operations introduce disturbances to the image that, while not visually apparent, significantly affect the image features in deep learning. These operations either introduce unrelated features or alter the original forgery features, making it difficult for the network to learn and detect forgery traces. After post-processing the image, PSCC-Net and HiFi-Net learn these forgery-independent features, which leads to performance degradation. In contrast, our proposed model effectively aggregates the RGB and noise features through FAM, which retains the  features well even through multiple convolutions, and effectively learns the local features of the image through ARPM, the combination of which leads to good robust performance of our model. 

\subsection{Ablation Study}
\label{subsec:ab}
\begin{table}[h]
	\setlength{\tabcolsep}{15pt}
	\caption{\textcolor{black}{Ablation Study Result}}
	\centering
	\renewcommand{\arraystretch}{1}
	\begin{tabularx}{\columnwidth}{@{}l*{2}{>{\centering\arraybackslash}X}@{}}
		\toprule
		Components & AUC &F1   \\
		\midrule
		RGB      & 76.8 &53.1   \\
		RGB + Bayar  & 79.1 &56.9   \\
		RGB + SRM  & 78.9 &57.1   \\
		RGB + GF  &79.2 &57.9  \\
		RGB + Guided Noise (GF+Sobel)& 80.8&57.1   \\
		RGB + Guided Noise + FAM  & \uline{85.4}&\uline{63.8}   \\
		RGB + GF + Bayar + FAM + ARPM& 85.8&55.7   \\
		RGB + GF + SRM + FAM + ARPM& 85.7&56.6   \\
		\midrule
		RGB + Guided Noise + FAM + ARPM (HRNet)  & 81.5&57.1  \\
		RGB + Guided Noise + FAM + ARPM (Ours)  & \textbf{89.4}&\textbf{66.5}   \\
		\bottomrule
	\end{tabularx}
	\label{AB_Study}
\end{table}
\vspace{-0.1cm}

To validate the effectiveness of our proposed various modules, we have separately tested the noise features, guided filter, Sobel filter, FAM and ARPM contributions to the model through extensive ablation experiments.

We sequentially add or remove our proposed modules on the model for ablation experiments and test them on the CASIA dataset. Table \ref{AB_Study} reports the results of the ablation experiments. From the Table \ref{AB_Study}, it can be seen that the model performs poorly when using only the RGB feature, and the performance improves with the introduction of the noise feature (RGB+GF), as well as the combination of the guided filter and Sobel Filters (RGB+Guided Noise). Compared to directly summing the RGB and noise features, the introduction of FAM (RGB+Guided Noise+FAM) to aggregate the features significantly improves the model performance, which is further improved by continuing to introduce ARPM to learn the local features. We also replace guided filter with BayarConv {\cite{BayarConv}} and SRM {\cite{SRM}} for experiments, respectively. The results show that the performance of all of them is not as good as using guided filter. In addition, we also replace the Sobel filter in our method using BayarConv and SRM, respectively, and the results show that the AUC after replacement is reduced by 3.6\% and 3.7\%, respectively, compared to using the Sobel filter. Therefore, all our proposed modules can effectively improve the model performance. In order to show the superiority of using EfficientNetV2 as the backbone network, we constructed a variant that replaces the backbone network with HRNet. From the table, it is easy to find that after replacing the backbone network with HRNet, the AUC and F1 scores of our method on the CASIA dataset decreased by 7.9\% and 9.4\%, respectively, which indicates that using EfficientNetV2 as the backbone network is superior to HRNet.

\subsection{Visualization}
\label{subsec:vis}
\subsubsection{Large Forged Regions Localization}
\begin{figure*}[t]
	\centering
	\includegraphics[width=5in]{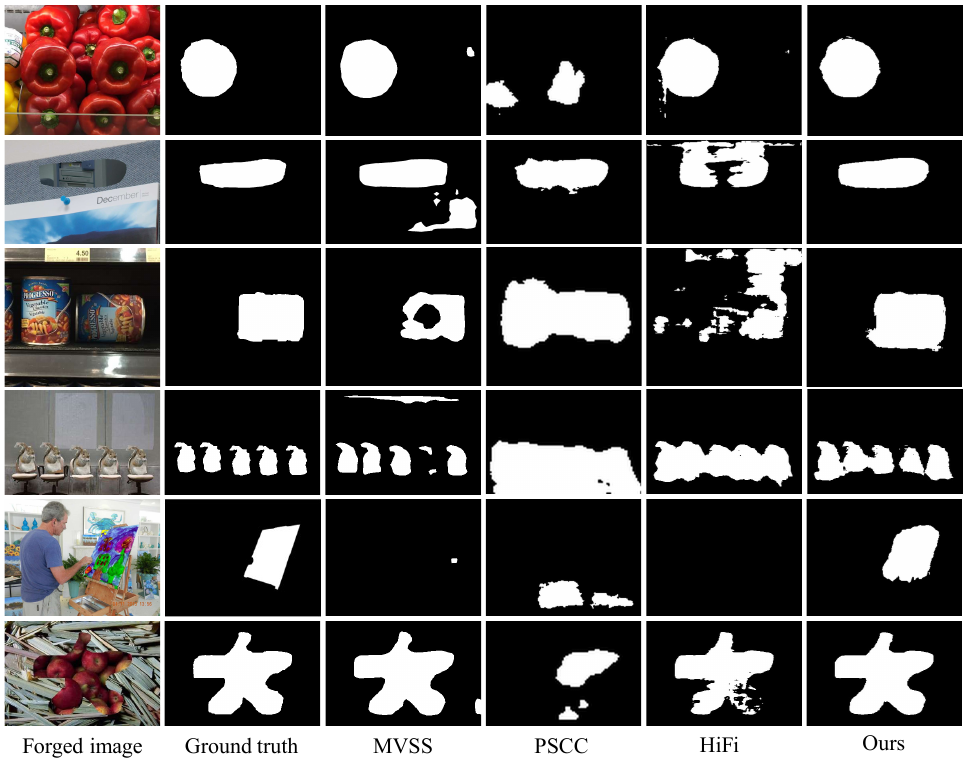}
	\caption{Examples of prediction masks for different methods on five public datasets. From left to right , we show forged images, ground-truth, predictions of MVSS-Net, PSCC-Net, HiFi-Net and the method we proposed.}
	\label{visual}
\end{figure*}
In order to visually represent the state-of-the-art of our model's localization performance, we describe the visualization results in this section. Fig. \ref{visual}  shows examples of localization results of our model with other state-of-the-art methods on five public datasets. Where for mask, 0 denote authentic pixels and 1 denotes forged pixels. Note that in order to distinguish small forged regions, the area of forged region is greater than 10\% for each image here. It can be seen that our model outperforms the other methods in terms of localization, with fewer cases of false alarms (e.g. rows 1, 3, 7 and 9). 

\subsubsection{Small Forged Regions Localization}
\begin{figure*}[!t]
	\centering
	\includegraphics[width=5in]{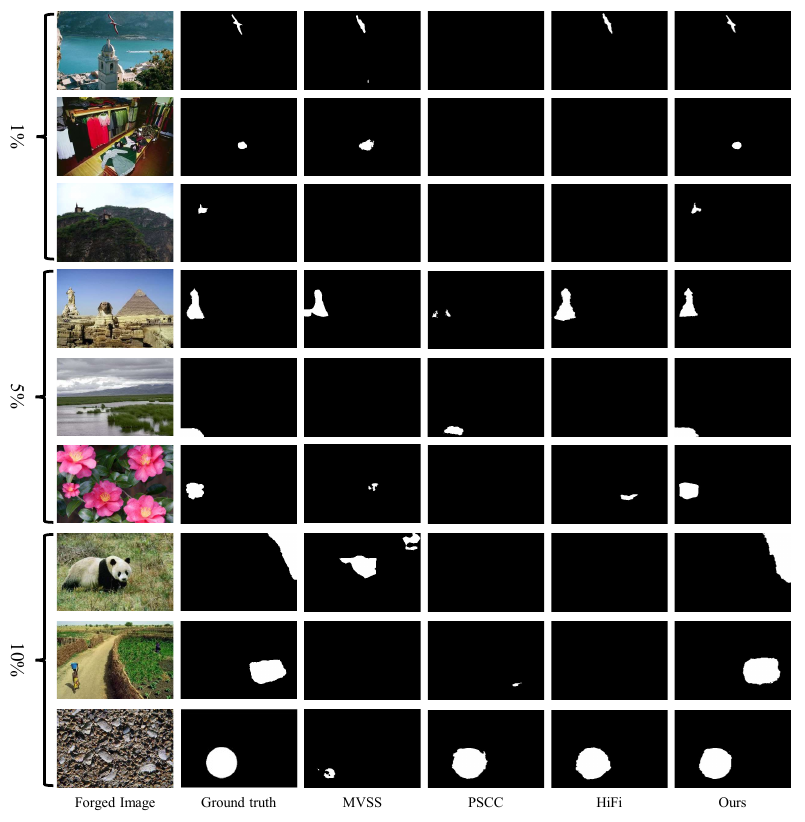}
	\caption{Examples of small forged regions localization for different methods. Most of the methods can not perform well in small area forgery detection and subject to false alarms or missed alarms.}
	\label{smallImage}
\end{figure*}
We have selected some examples from the three different area criteria chosen for the small forged regions experiments to demonstrate the performance of our model for localization in small forged areas, as shown in Fig. \ref{smallImage}. From the figure, we can easily see that, thanks to FAM and ARPM, our model also achieves good localization performance in small forged regions detection, and we can still successfully localize the forged region when the forged region is small, whereas some other methods fail to detect the forged region or incorrectly judge the real region as forged region. For example, in the first row of Fig. \ref{smallImage},  MVSS-Net roughly identifies the forged region but produces a false alarm at the bottom of the image, while PSCC-Net fails to detect the forged area, resulting in a missed detection. HiFi-Net slightly improves by outlining the forged region in the shape of a bird, but our method accurately detects and delineates the forged region with precision.

\section{Conclusion}
\label{conclusion}
In this article, we propose a new end-to-end IFL network structure. We apply guided filter in combination with Sobel filter for noise feature extraction to enable our model to acquire forgery traces of different forgery types while enhancing the edge artifact. We propose an FAM to better aggregate RGB and noise features to reduce feature degradation, and ARPM to enable our model to efficiently learn the global and local features of the image to enhance the feature representation. Extensive experiments  demonstrate that the performance of our model outperforms the current state-of-the-art methods, while the robustness of our model is equally superior to other IFL methods due to the effective learning of forgery features by our proposed FAM and ARPM.

\section{CRediT authorship contribution statement}
\textbf{Yakun Niu}:Conceptualization, Writing - Review \& Editing, Resources, Funding acquisition. \textbf{Pei Chen}:Writing - Original Draft, Methodology, Software, Data Curation, Formal analysis \textbf{Lei Zhang}:Supervision, Project administration, Funding acquisition. \textbf{Lei Tan}:Validation, Conceptualization. \textbf{Yingjian Chen}:Investigation, Visualization.

\section{Declaration of competing interest}
The authors declare that they have no known competing financial interests or personal relationships that could have appeared to
influence the work reported in this paper.

\section{Data availability}
Data will be made available on request.

\section{Acknowledgments}
This work was supported in part by the National Natural Science Foundation of China (Grant 62202141), in part by Henan Province Science and Technology Research Project (Grant 232102240020), in part by University Young Key Teacher of Henan Province.(Grant 2020GGJS027)

\bibliography{ref}

\begin{thebibliography}{10}
\expandafter\ifx\csname url\endcsname\relax
  \def\url#1{\texttt{#1}}\fi
\expandafter\ifx\csname urlprefix\endcsname\relax\def\urlprefix{URL }\fi
\expandafter\ifx\csname href\endcsname\relax
  \def\href#1#2{#2} \def\path#1{#1}\fi

\bibitem{blur}
I.~T. Young, L.~J. {van Vliet}, {Recursive implementation of the Gaussian
  filter}, Signal Processing 44~(2) (1995) 139--151.
\newblock \href {https://doi.org/https://doi.org/10.1016/0165-1684(95)00020-E}
  {\path{doi:https://doi.org/10.1016/0165-1684(95)00020-E}}.

\bibitem{Splice}
M.~Huh, A.~Liu, A.~Owens, A.~A. Efros, {Fighting Fake News: Image Splice
  Detection via Learned Self-Consistency}, in: Proceedings of the European
  Conference on Computer Vision (ECCV), 2018.
\newblock \href {https://doi.org/https://doi.org/10.1007/978-3-030-01252-6_7}
  {\path{doi:https://doi.org/10.1007/978-3-030-01252-6_7}}.

\bibitem{CopyMove1}
G.~Gani, F.~Qadir, {A robust copy-move forgery detection technique based on
  discrete cosine transform and cellular automata}, Journal of Information
  Security and Applications 54 (2020) 102510.
\newblock \href {https://doi.org/https://doi.org/10.1016/j.jisa.2020.102510}
  {\path{doi:https://doi.org/10.1016/j.jisa.2020.102510}}.

\bibitem{PassiveForgery}
W.~F. Mashaan, I.~T. Ahmed, {Passive Forgery Detection Techniques:A Survey},
  in: 2023 IEEE International Conference on Automatic Control and Intelligent
  Systems (I2CACIS), 2023, pp. 321--326.
\newblock \href {https://doi.org/10.1109/I2CACIS57635.2023.10193581}
  {\path{doi:10.1109/I2CACIS57635.2023.10193581}}.

\bibitem{GAN}
M.~Krichen, {Generative Adversarial Networks}, in: 2023 14th International
  Conference on Computing Communication and Networking Technologies (ICCCNT),
  2023, pp. 1--7.
\newblock \href {https://doi.org/10.1109/ICCCNT56998.2023.10306417}
  {\path{doi:10.1109/ICCCNT56998.2023.10306417}}.

\bibitem{MVSS}
C.~Dong, X.~Chen, R.~Hu, J.~Cao, X.~Li, {MVSS-Net: Multi-View Multi-Scale
  Supervised Networks for Image Manipulation Detection}, IEEE Transactions on
  Pattern Analysis and Machine Intelligence 45~(3) (2023) 3539--3553.
\newblock \href {https://doi.org/10.1109/TPAMI.2022.3180556}
  {\path{doi:10.1109/TPAMI.2022.3180556}}.

\bibitem{PSCC}
X.~Liu, Y.~Liu, J.~Chen, X.~Liu, {PSCC-Net: Progressive Spatio-Channel
  Correlation Network for Image Manipulation Detection and Localization}, IEEE
  Transactions on Circuits and Systems for Video Technology 32~(11) (2022)
  7505--7517.
\newblock \href {https://doi.org/10.1109/TCSVT.2022.3189545}
  {\path{doi:10.1109/TCSVT.2022.3189545}}.

\bibitem{HiFi}
X.~Guo, X.~Liu, Z.~Ren, S.~Grosz, I.~Masi, X.~Liu, {Hierarchical Fine-Grained
  Image Forgery Detection and Localization}, in: 2023 IEEE/CVF Conference on
  Computer Vision and Pattern Recognition (CVPR), 2023, pp. 3155--3165.
\newblock \href {https://doi.org/10.1109/CVPR52729.2023.00308}
  {\path{doi:10.1109/CVPR52729.2023.00308}}.

\bibitem{GF1}
K.~He, J.~Sun, X.~Tang, {Guided Image Filtering}, IEEE Transactions on Pattern
  Analysis and Machine Intelligence 35~(6) (2013) 1397--1409.
\newblock \href {https://doi.org/10.1109/TPAMI.2012.213}
  {\path{doi:10.1109/TPAMI.2012.213}}.

\bibitem{Sobel}
R.~E. Twogood, F.~G. Sommer, {Digital Image Processing}, IEEE Transactions on
  Nuclear Science 29~(3) (1982) 1075--1086.
\newblock \href {https://doi.org/10.1109/TNS.1982.4336327}
  {\path{doi:10.1109/TNS.1982.4336327}}.

\bibitem{EffV2}
M.~Tan, Q.~Le, {EfficientNetV2: Smaller Models and Faster Training}, in:
  M.~Meila, T.~Zhang (Eds.), Proceedings of the 38th International Conference
  on Machine Learning, Vol. 139 of Proceedings of Machine Learning Research,
  PMLR, 2021, pp. 10096--10106.
\newblock \href {https://doi.org/https://doi.org/10.48550/arXiv.2104.00298}
  {\path{doi:https://doi.org/10.48550/arXiv.2104.00298}}.

\bibitem{DC}
Y.~Chen, X.~Dai, M.~Liu, D.~Chen, L.~Yuan, Z.~Liu, {Dynamic Convolution:
  Attention Over Convolution Kernels}, in: 2020 IEEE/CVF Conference on Computer
  Vision and Pattern Recognition (CVPR), 2020, pp. 11027--11036.
\newblock \href {https://doi.org/10.1109/CVPR42600.2020.01104}
  {\path{doi:10.1109/CVPR42600.2020.01104}}.

\bibitem{Atrous}
F.~Yu, V.~Koltun, {Multi-Scale Context Aggregation by Dilated Convolutions},
  in: International Conference on Learning Representations (ICLR), 2016.
\newblock \href {https://doi.org/https://doi.org/10.48550/arXiv.1511.07122}
  {\path{doi:https://doi.org/10.48550/arXiv.1511.07122}}.

\bibitem{ResNet50}
K.~He, X.~Zhang, S.~Ren, J.~Sun, {Deep Residual Learning for Image
  Recognition}, in: 2016 IEEE Conference on Computer Vision and Pattern
  Recognition (CVPR), 2016, pp. 770--778.
\newblock \href {https://doi.org/10.1109/CVPR.2016.90}
  {\path{doi:10.1109/CVPR.2016.90}}.

\bibitem{Attention}
A.~Vaswani, N.~Shazeer, N.~Parmar, J.~Uszkoreit, L.~Jones, A.~N. Gomez,
  L.~Kaiser, I.~Polosukhin, {Attention is all you need}, in: Proceedings of the
  31st International Conference on Neural Information Processing Systems,
  NIPS'17, Curran Associates Inc., Red Hook, NY, USA, 2017, p. 6000–6010.
\newblock \href {https://doi.org/https://doi.org/10.48550/arXiv.1706.03762}
  {\path{doi:https://doi.org/10.48550/arXiv.1706.03762}}.

\bibitem{JLSTM}
J.~H. Bappy, A.~K. Roy-Chowdhury, J.~Bunk, L.~Nataraj, B.~Manjunath,
  {Exploiting Spatial Structure for Localizing Manipulated Image Regions}, in:
  2017 IEEE International Conference on Computer Vision (ICCV), 2017, pp.
  4980--4989.
\newblock \href {https://doi.org/10.1109/ICCV.2017.532}
  {\path{doi:10.1109/ICCV.2017.532}}.

\bibitem{RGBNet}
P.~Zhou, X.~Han, V.~I. Morariu, L.~S. Davis, {Learning Rich Features for Image
  Manipulation Detection}, in: 2018 IEEE/CVF Conference on Computer Vision and
  Pattern Recognition, 2018, pp. 1053--1061.
\newblock \href {https://doi.org/10.1109/CVPR.2018.00116}
  {\path{doi:10.1109/CVPR.2018.00116}}.

\bibitem{HLSTM}
J.~H. Bappy, C.~Simons, L.~Nataraj, B.~S. Manjunath, A.~K. Roy-Chowdhury,
  {Hybrid LSTM and Encoder–Decoder Architecture for Detection of Image
  Forgeries}, IEEE Transactions on Image Processing 28~(7) (2019) 3286--3300.
\newblock \href {https://doi.org/10.1109/TIP.2019.2895466}
  {\path{doi:10.1109/TIP.2019.2895466}}.

\bibitem{ManTra}
Y.~Wu, W.~AbdAlmageed, P.~Natarajan, {ManTra-Net: Manipulation Tracing Network
  for Detection and Localization of Image Forgeries With Anomalous Features},
  in: 2019 IEEE/CVF Conference on Computer Vision and Pattern Recognition
  (CVPR), 2019, pp. 9535--9544.
\newblock \href {https://doi.org/10.1109/CVPR.2019.00977}
  {\path{doi:10.1109/CVPR.2019.00977}}.

\bibitem{SPAN}
X.~Hu, Z.~Zhang, Z.~Jiang, S.~Chaudhuri, Z.~Yang, R.~Nevatia, {SPAN: Spatial
  Pyramid Attention Network for Image Manipulation Localization}, in:
  A.~Vedaldi, H.~Bischof, T.~Brox, J.-M. Frahm (Eds.), Computer Vision -- ECCV
  2020, Springer International Publishing, Cham, 2020, pp. 312--328.
\newblock \href {https://doi.org/https://doi.org/10.1007/978-3-030-58589-1_19}
  {\path{doi:https://doi.org/10.1007/978-3-030-58589-1_19}}.

\bibitem{CAT}
M.-J. Kwon, I.-J. Yu, S.-H. Nam, H.-K. Lee, {CAT-Net: Compression Artifact
  Tracing Network for Detection and Localization of Image Splicing}, in: 2021
  IEEE Winter Conference on Applications of Computer Vision (WACV), 2021, pp.
  375--384.
\newblock \href {https://doi.org/10.1109/WACV48630.2021.00042}
  {\path{doi:10.1109/WACV48630.2021.00042}}.

\bibitem{EMTNet}
X.~Lin, S.~Wang, J.~Deng, Y.~Fu, X.~Bai, X.~Chen, X.~Qu, W.~Tang, {Image
  manipulation detection by multiple tampering traces and edge artifact
  enhancement}, Pattern Recognition 133 (2023) 109026.
\newblock \href {https://doi.org/https://doi.org/10.1016/j.patcog.2022.109026}
  {\path{doi:https://doi.org/10.1016/j.patcog.2022.109026}}.

\bibitem{EVP}
W.~Liu, X.~Shen, C.-M. Pun, X.~Cun, {Explicit Visual Prompting for Low-Level
  Structure Segmentations}, in: 2023 IEEE/CVF Conference on Computer Vision and
  Pattern Recognition (CVPR), 2023, pp. 19434--19445.
\newblock \href {https://doi.org/10.1109/CVPR52729.2023.01862}
  {\path{doi:10.1109/CVPR52729.2023.01862}}.

\bibitem{Edgebased}
Y.~Zhou, H.~Wang, Q.~Zeng, R.~Zhang, S.~Meng, {Exploring weakly-supervised
  image manipulation localization with tampering Edge-based class activation
  map}, Expert Systems with Applications 249 (2024) 123501.
\newblock \href {https://doi.org/https://doi.org/10.1016/j.eswa.2024.123501}
  {\path{doi:https://doi.org/10.1016/j.eswa.2024.123501}}.

\bibitem{DCT}
Z.~Lin, J.~He, X.~Tang, C.-K. Tang, {Fast, automatic and fine-grained tampered
  JPEG image detection via DCT coefficient analysis}, Pattern Recognition
  42~(11) (2009) 2492--2501.
\newblock \href {https://doi.org/https://doi.org/10.1016/j.patcog.2009.03.019}
  {\path{doi:https://doi.org/10.1016/j.patcog.2009.03.019}}.

\bibitem{CFA}
P.~Ferrara, T.~Bianchi, A.~De~Rosa, A.~Piva, {Image Forgery Localization via
  Fine-Grained Analysis of CFA Artifacts}, IEEE Transactions on Information
  Forensics and Security 7~(5) (2012) 1566--1577.
\newblock \href {https://doi.org/10.1109/TIFS.2012.2202227}
  {\path{doi:10.1109/TIFS.2012.2202227}}.

\bibitem{Splicebuster}
D.~Cozzolino, G.~Poggi, L.~Verdoliva, {Splicebuster: A new blind image splicing
  detector}, in: 2015 IEEE International Workshop on Information Forensics and
  Security (WIFS), 2015, pp. 1--6.
\newblock \href {https://doi.org/10.1109/WIFS.2015.7368565}
  {\path{doi:10.1109/WIFS.2015.7368565}}.

\bibitem{Copy-move_Tradition}
D.~Vaishnavi, T.~Subashini, {Application of local invariant symmetry features
  to detect and localize image copy move forgeries}, Journal of Information
  Security and Applications 44 (2019) 23--31.
\newblock \href {https://doi.org/https://doi.org/10.1016/j.jisa.2018.11.001}
  {\path{doi:https://doi.org/10.1016/j.jisa.2018.11.001}}.

\bibitem{Tetrolet}
K.~B. Meena, V.~Tyagi, {A copy-move image forgery detection technique based on
  tetrolet transform}, Journal of Information Security and Applications 52
  (2020) 102481.
\newblock \href {https://doi.org/https://doi.org/10.1016/j.jisa.2020.102481}
  {\path{doi:https://doi.org/10.1016/j.jisa.2020.102481}}.

\bibitem{SURF}
B.~Soni, P.~K. Das, D.~M. Thounaojam, {Geometric transformation invariant block
  based copy-move forgery detection using fast and efficient hybrid local
  features}, Journal of Information Security and Applications 45 (2019) 44--51.
\newblock \href {https://doi.org/https://doi.org/10.1016/j.jisa.2019.01.007}
  {\path{doi:https://doi.org/10.1016/j.jisa.2019.01.007}}.

\bibitem{CNN}
Y.~Lecun, L.~Bottou, Y.~Bengio, P.~Haffner, {Gradient-based learning applied to
  document recognition}, Proceedings of the IEEE 86~(11) (1998) 2278--2324.
\newblock \href {https://doi.org/10.1109/5.726791}
  {\path{doi:10.1109/5.726791}}.

\bibitem{FasterRCNN}
S.~Ren, K.~He, R.~Girshick, J.~Sun, {Faster R-CNN: Towards Real-Time Object
  Detection with Region Proposal Networks}, IEEE Transactions on Pattern
  Analysis and Machine Intelligence 39~(6) (2017) 1137--1149.
\newblock \href {https://doi.org/10.1109/TPAMI.2016.2577031}
  {\path{doi:10.1109/TPAMI.2016.2577031}}.

\bibitem{SRM}
J.~Fridrich, J.~Kodovsky, {Rich Models for Steganalysis of Digital Images},
  IEEE Transactions on Information Forensics and Security 7~(3) (2012)
  868--882.
\newblock \href {https://doi.org/10.1109/TIFS.2012.2190402}
  {\path{doi:10.1109/TIFS.2012.2190402}}.

\bibitem{BayarConv}
B.~Bayar, M.~C. Stamm, {Constrained Convolutional Neural Networks: A New
  Approach Towards General Purpose Image Manipulation Detection}, IEEE
  Transactions on Information Forensics and Security 13~(11) (2018) 2691--2706.
\newblock \href {https://doi.org/10.1109/TIFS.2018.2825953}
  {\path{doi:10.1109/TIFS.2018.2825953}}.

\bibitem{HRNet}
J.~Wang, K.~Sun, T.~Cheng, B.~Jiang, C.~Deng, Y.~Zhao, D.~Liu, Y.~Mu, M.~Tan,
  X.~Wang, W.~Liu, B.~Xiao, {Deep High-Resolution Representation Learning for
  Visual Recognition}, IEEE Transactions on Pattern Analysis and Machine
  Intelligence 43~(10) (2021) 3349--3364.
\newblock \href {https://doi.org/10.1109/TPAMI.2020.2983686}
  {\path{doi:10.1109/TPAMI.2020.2983686}}.

\bibitem{SVM}
S.~Bharathiraja, B.~R. Kanna, S.~Geetha, M.~Hariharan, {Exposing digital image
  forgeries from statistical footprints}, Journal of Information Security and
  Applications 69 (2022) 103273.
\newblock \href {https://doi.org/https://doi.org/10.1016/j.jisa.2022.103273}
  {\path{doi:https://doi.org/10.1016/j.jisa.2022.103273}}.

\bibitem{NoisePrint}
D.~Cozzolino, L.~Verdoliva, {Noiseprint: A CNN-Based Camera Model Fingerprint},
  IEEE Transactions on Information Forensics and Security 15 (2020) 144--159.
\newblock \href {https://doi.org/10.1109/TIFS.2019.2916364}
  {\path{doi:10.1109/TIFS.2019.2916364}}.

\bibitem{TurFor}
F.~Guillaro, D.~Cozzolino, A.~Sud, N.~Dufour, L.~Verdoliva, {TruFor: Leveraging
  All-Round Clues for Trustworthy Image Forgery Detection and Localization},
  in: 2023 IEEE/CVF Conference on Computer Vision and Pattern Recognition
  (CVPR), 2023, pp. 20606--20615.
\newblock \href {https://doi.org/10.1109/CVPR52729.2023.01974}
  {\path{doi:10.1109/CVPR52729.2023.01974}}.

\bibitem{CMX}
J.~Zhang, H.~Liu, K.~Yang, X.~Hu, R.~Liu, R.~Stiefelhagen, {CMX: Cross-Modal
  Fusion for RGB-X Semantic Segmentation With Transformers}, IEEE Transactions
  on Intelligent Transportation Systems 24~(12) (2023) 14679--14694.
\newblock \href {https://doi.org/10.1109/TITS.2023.3300537}
  {\path{doi:10.1109/TITS.2023.3300537}}.

\bibitem{DFMM}
X.~Xia, L.~C. Su, S.~P. Wang, X.~Y. Li, {DMFF-Net: Double-stream multilevel
  feature fusion network for image forgery localization}, Engineering
  Applications of Artificial Intelligence 127 (2024) 107200.
\newblock \href
  {https://doi.org/https://doi.org/10.1016/j.engappai.2023.107200}
  {\path{doi:https://doi.org/10.1016/j.engappai.2023.107200}}.

\bibitem{AFTLNet}
X.~Ding, Y.~Deng, Y.~Zhao, W.~Zhu, {AFTLNet: An efficient adaptive forgery
  traces learning network for deep image inpainting localization}, Journal of
  Information Security and Applications 84 (2024) 103825.
\newblock \href {https://doi.org/https://doi.org/10.1016/j.jisa.2024.103825}
  {\path{doi:https://doi.org/10.1016/j.jisa.2024.103825}}.

\bibitem{LSTM}
S.~Hochreiter, J.~Schmidhuber, {Long Short-Term Memory}, Neural Computation
  9~(8) (1997) 1735--1780.
\newblock \href {https://doi.org/10.1162/neco.1997.9.8.1735}
  {\path{doi:10.1162/neco.1997.9.8.1735}}.

\bibitem{PLGNet}
Z.~Shi, X.~Shen, H.~Chen, Y.~Lyu, {PL-GNet: Pixel Level Global Network for
  detection and localization of image forgeries}, Signal Processing: Image
  Communication 119 (2023) 117029.
\newblock \href {https://doi.org/https://doi.org/10.1016/j.image.2023.117029}
  {\path{doi:https://doi.org/10.1016/j.image.2023.117029}}.

\bibitem{SCFE}
F.~Li, H.~Zhai, X.~Zhang, C.~Qin, {Image Manipulation Localization Using
  Spatial–Channel Fusion Excitation and Fine-Grained Feature Enhancement},
  IEEE Transactions on Instrumentation and Measurement 73 (2024) 1--14.
\newblock \href {https://doi.org/10.1109/TIM.2023.3338703}
  {\path{doi:10.1109/TIM.2023.3338703}}.

\bibitem{Eff}
M.~Tan, Q.~Le, {{E}fficient{N}et: Rethinking Model Scaling for Convolutional
  Neural Networks}, in: K.~Chaudhuri, R.~Salakhutdinov (Eds.), Proceedings of
  the 36th International Conference on Machine Learning, Vol.~97 of Proceedings
  of Machine Learning Research, PMLR, 2019, pp. 6105--6114.
\newblock \href {https://doi.org/https://doi.org/10.48550/arXiv.1905.11946}
  {\path{doi:https://doi.org/10.48550/arXiv.1905.11946}}.

\bibitem{SNIS}
J.~Chen, X.~Liao, W.~Wang, Z.~Qian, Z.~Qin, Y.~Wang, {SNIS: A Signal Noise
  Separation-Based Network for Post-Processed Image Forgery Detection}, IEEE
  Transactions on Circuits and Systems for Video Technology 33~(2) (2023)
  935--951.
\newblock \href {https://doi.org/10.1109/TCSVT.2022.3204753}
  {\path{doi:10.1109/TCSVT.2022.3204753}}.

\bibitem{GF}
Z.~Guo, G.~Yang, J.~Chen, X.~Sun, {Exposing Deepfake Face Forgeries With Guided
  Residuals}, IEEE Transactions on Multimedia 25 (2023) 8458--8470.
\newblock \href {https://doi.org/10.1109/TMM.2023.3237169}
  {\path{doi:10.1109/TMM.2023.3237169}}.

\bibitem{PRNU}
G.~Chierchia, D.~Cozzolino, G.~Poggi, C.~Sansone, L.~Verdoliva, {Guided
  filtering for PRNU-based localization of small-size image forgeries}, in:
  2014 IEEE International Conference on Acoustics, Speech and Signal Processing
  (ICASSP), 2014, pp. 6231--6235.
\newblock \href {https://doi.org/10.1109/ICASSP.2014.6854802}
  {\path{doi:10.1109/ICASSP.2014.6854802}}.

\bibitem{FPN}
T.-Y. Lin, P.~Dollár, R.~Girshick, K.~He, B.~Hariharan, S.~Belongie, {Feature
  Pyramid Networks for Object Detection}, in: 2017 IEEE Conference on Computer
  Vision and Pattern Recognition (CVPR), 2017, pp. 936--944.
\newblock \href {https://doi.org/10.1109/CVPR.2017.106}
  {\path{doi:10.1109/CVPR.2017.106}}.

\bibitem{NonLocal}
X.~Wang, R.~Girshick, A.~Gupta, K.~He, {Non-local Neural Networks}, in: 2018
  IEEE/CVF Conference on Computer Vision and Pattern Recognition, 2018, pp.
  7794--7803.
\newblock \href {https://doi.org/10.1109/CVPR.2018.00813}
  {\path{doi:10.1109/CVPR.2018.00813}}.

\bibitem{SENet}
J.~Hu, L.~Shen, G.~Sun, {Squeeze-and-Excitation Networks}, in: 2018 IEEE/CVF
  Conference on Computer Vision and Pattern Recognition, 2018, pp. 7132--7141.
\newblock \href {https://doi.org/10.1109/CVPR.2018.00745}
  {\path{doi:10.1109/CVPR.2018.00745}}.

\bibitem{MSCOCO}
T.-Y. Lin, M.~Maire, S.~Belongie, J.~Hays, P.~Perona, D.~Ramanan,
  P.~Doll{\'a}r, C.~L. Zitnick, {Microsoft COCO: Common Objects in Context},
  in: D.~Fleet, T.~Pajdla, B.~Schiele, T.~Tuytelaars (Eds.), Computer Vision --
  ECCV 2014, Springer International Publishing, Cham, 2014, pp. 740--755.
\newblock \href {https://doi.org/https://doi.org/10.1007/978-3-319-10602-1_48}
  {\path{doi:https://doi.org/10.1007/978-3-319-10602-1_48}}.

\bibitem{CopyMove}
Y.~Wu, W.~Abd-Almageed, P.~Natarajan, {BusterNet: Detecting Copy-Move Image
  Forgery with Source/Target Localization}, in: V.~Ferrari, M.~Hebert,
  C.~Sminchisescu, Y.~Weiss (Eds.), Computer Vision -- ECCV 2018, Springer
  International Publishing, Cham, 2018, pp. 170--186.
\newblock \href {https://doi.org/https://doi.org/10.1007/978-3-030-01231-1_11}
  {\path{doi:https://doi.org/10.1007/978-3-030-01231-1_11}}.

\bibitem{Columbia}
Y.-F. Hsu, S.-F. Chang, {Detecting Image Splicing Using Geometry Invariants and
  Camera Characteristics Consistency}, in: International Conference on
  Multimedia and Expo, 2006.
\newblock \href {https://doi.org/10.1109/ICME.2006.262447}
  {\path{doi:10.1109/ICME.2006.262447}}.

\bibitem{Coverage}
B.~Wen, Y.~Zhu, R.~Subramanian, T.-T. Ng, X.~Shen, S.~Winkler, {COVERAGE — A
  novel database for copy-move forgery detection}, in: 2016 IEEE International
  Conference on Image Processing (ICIP), 2016, pp. 161--165.
\newblock \href {https://doi.org/10.1109/ICIP.2016.7532339}
  {\path{doi:10.1109/ICIP.2016.7532339}}.

\bibitem{CASIA}
J.~Dong, W.~Wang, T.~Tan, {CASIA Image Tampering Detection Evaluation
  Database}, in: 2013 IEEE China Summit and International Conference on Signal
  and Information Processing, 2013, pp. 422--426.
\newblock \href {https://doi.org/10.1109/ChinaSIP.2013.6625374}
  {\path{doi:10.1109/ChinaSIP.2013.6625374}}.

\bibitem{NIST16}
H.~Guan, M.~Kozak, E.~Robertson, Y.~Lee, A.~N. Yates, A.~Delgado, D.~Zhou,
  T.~Kheyrkhah, J.~Smith, J.~Fiscus, {MFC Datasets: Large-Scale Benchmark
  Datasets for Media Forensic Challenge Evaluation}, in: 2019 IEEE Winter
  Applications of Computer Vision Workshops (WACVW), 2019, pp. 63--72.
\newblock \href {https://doi.org/10.1109/WACVW.2019.00018}
  {\path{doi:10.1109/WACVW.2019.00018}}.

\bibitem{IMD20}
A.~Novozámský, B.~Mahdian, S.~Saic, {IMD2020: A Large-Scale Annotated Dataset
  Tailored for Detecting Manipulated Images}, in: 2020 IEEE Winter Applications
  of Computer Vision Workshops (WACVW), 2020, pp. 71--80.
\newblock \href {https://doi.org/10.1109/WACVW50321.2020.9096940}
  {\path{doi:10.1109/WACVW50321.2020.9096940}}.

\bibitem{Adam}
D.~P. Kingma, J.~Ba, {Adam: A Method for Stochastic Optimization}, CoRR
  abs/1412.6980 (2014).
\newblock \href {https://doi.org/https://doi.org/10.48550/arXiv.1412.6980}
  {\path{doi:https://doi.org/10.48550/arXiv.1412.6980}}.

\bibitem{EITLNet}
K.~Guo, H.~Zhu, G.~Cao, {Effective Image Tampering Localization Via Enhanced
  Transformer and Co-Attention Fusion}, in: ICASSP 2024 - 2024 IEEE
  International Conference on Acoustics, Speech and Signal Processing (ICASSP),
  2024, pp. 4895--4899.
\newblock \href {https://doi.org/10.1109/ICASSP48485.2024.10446332}
  {\path{doi:10.1109/ICASSP48485.2024.10446332}}.

\bibitem{ObjectFormer}
J.~Wang, Z.~Wu, J.~Chen, X.~Han, A.~Shrivastava, S.-N. Lim, Y.-G. Jiang,
  {ObjectFormer for Image Manipulation Detection and Localization}, in: 2022
  IEEE/CVF Conference on Computer Vision and Pattern Recognition (CVPR), 2022,
  pp. 2354--2363.
\newblock \href {https://doi.org/10.1109/CVPR52688.2022.00240}
  {\path{doi:10.1109/CVPR52688.2022.00240}}.

\end{thebibliography}
\end{document}